\documentclass[jmse,article,accept,pdftex,moreauthors]{Definitions/mdpi} 

\firstpage{1} 
\pubvolume{10}
\issuenum{9}
\articlenumber{1289}
\pubyear{2022}
\copyrightyear{2022}
\externaleditor{Academic Editor: {Antoni Burguera}}
\datereceived{19 August 2022} 
\dateaccepted{7 September 2022} 
\datepublished{13 September 2022} 
\hreflink{\textls[-35]{https://doi.org/10.3390/jmse10091289}} 
\pdfoutput=1

 \usetikzlibrary{matrix}
\usepackage{amssymb}
\usepackage{amsmath}
\usepackage{tikz}
\usepackage{pgfplots}

\usepackage{booktabs}
\usepackage{multirow}
\usepackage{graphicx}
\usepackage{lscape}
\pgfplotsset{compat=newest}
\usepgfplotslibrary{units}

 \DeclareMathAlphabet{\pazocal}{OMS}{zplm}{m}{n}
\usepackage[]{trackchanges}
\addeditor{IP}

\Title{Virtual Underwater Datasets for Autonomous Inspections}

\TitleCitation{Virtual Underwater Datasets for Autonomous Inspections}

\Author{{Ioannis} 
 Polymenis \orcidA{}, Maryam Haroutunian, Rose Norman *\orcidB{} and David Trodden}

\AuthorNames{Ioannis Polymenis, Maryam Haroutunian, Rose Norman and David Trodden}

\AuthorCitation{Polymenis, I.; Haroutunian,~M.; Norman, R.; Trodden, D.}

\address[1]{School of Engineering, Newcastle University, Newcastle upon Tyne NE1 7RU, UK
}

\corres{\hangafter=1 \hangindent=1.05em \hspace{-0.82em} Correspondence: {rose.norman@ncl.ac.uk} 
}

\abstract{Underwater Vehicles have become more sophisticated, driven by the off-shore sector and the scientific community's rapid advancements in underwater operations. Notably, many underwater tasks, including the assessment of subsea infrastructure, are performed with the assistance of Autonomous Underwater Vehicles (AUVs). There have been recent breakthroughs in Artificial Intelligence (AI) and, notably, Deep Learning (DL) models and applications, which have widespread usage in a variety of fields, including aerial unmanned vehicles, autonomous car navigation, and other applications. However, they are not as prevalent in underwater applications due to the difficulty of obtaining underwater datasets for a specific application. In this sense, the current study utilises recent advancements in the area of DL to construct a bespoke dataset generated from photographs of items captured in a laboratory environment. Generative Adversarial Networks (GANs) were utilised to translate the laboratory object dataset into the underwater domain by combining the collected images with photographs containing the underwater environment. The findings demonstrated the feasibility of creating such a dataset, since the resulting images closely resembled the real underwater environment when compared with real-world underwater ship hull images. Therefore, the artificial datasets of the underwater environment can overcome the difficulties arising from the limited access to real-world underwater images and are used to enhance underwater operations through underwater object image classification and detection.  
}

\begin{document}


\vspace{6pt} 

\section{Introduction}
Remotely Operated Vehicles (ROVs) and Autonomous Underwater Vehicles (AUVs) are used extensively by the offshore oil and gas industry and the offshore renewables sector, as well as by the marine scientific community, to carry out tasks such as the inspection and maintenance of underwater structures and the survey of the oceans in a manner that is both safe and accurate. To a significant extent, missions carried out by ROVs and AUVs depend on visual inputs to accurately portray and comprehend the surrounding subsea world. As a consequence of this, current underwater intervention requires the use of
image classification and object detection techniques. ROVs have been developed and optimised in terms of underwater operations and object manipulations to develop robust vehicles that make physical intervention possible. This was a result of research needs as well as industry requirements for more complex and demanding operations. The same industrial and scientific needs have resulted in the optimisation of AUVs, specifically in terms of power requirements, manoeuvrability, navigation, communication, and autonomy. This establishes AUVs as the predominant means for underwater operations in areas that are hazardous for humans to be in~\cite{Ridolfi2015}.

As a consequence of this, autonomous underwater intervention has to make use of the most promising available technologies of Artificial Intelligence (AI), namely those of Deep Learning (DL) and Computer Vision (CV).

The existing research on autonomous underwater operations takes advantage of advancements in DL and CV and offers practical solutions to the issue of underwater image enhancement and restoration by increasing the resolution and contrast of underwater images~\cite{Corchs2010}. In addition, the work of~\cite{Han2020} improves the image quality by resolving the issue of poor lighting that occurs in an underwater setting. 
Although these methods provide outcomes that are considered to be state of the art, none of the existing methods deals with the challenge of having access to readily available underwater image datasets.

Currently, the problem of readily available datasets is well known to research areas from Unmanned Aerial Vehicles (UAVs) applications~\cite{cazzatoSurvey2020a} to unmanned underwater operations~\cite{Chen2020},  and it is the fundamental issue that needs to be addressed to improve the performance of the different learning and detection models used by those fields. Particularly, for underwater operations, the issue arises primarily because of the high cost and the difficulties that need to be overcome to collect such a dataset from real-world underwater missions~\cite{Chen2020}. Because of this, the current work addresses the issues mentioned above and creates a model that enables the generation of custom underwater images by making use of common objects that can be found in underwater structures such as gas pipelines, underwater cables, oil/gas wells, wind turbine piles, etc. In addition, images with characteristics similar to those seen in the natural subsea environment were produced using recent developments in DL models.

\subsection*{Related Work}
Unmanned robots need to understand their complex environment to achieve complete autonomous operational capabilities, with object detection being the fundamental low-level task~\cite{cazzatoSurvey2020a}. The demand for underwater vehicles to achieve autonomous capabilities is even more demanding due to the challenging environmental conditions. 
Underwater Vehicles are equipped with various sensors and instrumentation such as GPS, cameras, LiDAR cameras, and sonars~\cite{xuUnderwaterVisualNavigation2021, yangUnderwaterPositioningSystem2022}. Cameras are essential because they allow for visual interaction between the user/operator and the vehicle~\cite{ludvigsenIntegratedAutonomousUnderwater2016}.

During the past few years, object detection models have become more sophisticated and accurate than ever before, and they are able to take advantage of the modern embedded systems~\cite{cazzatoSurvey2020a}.
Some landmark architectures that revolutionise modern CV applications include the R-CNN family of models~\cite{RichFeatureHierarchies2014b, girshickFastRCNN2015, heMaskRCNN2018}, the YOLO architecture and its different \mbox{versions~\cite{Redmon2016b, bochkovskiyYoloV4V32021}}, and Feature Pyramid Networks (FPN)~\cite{cazzatoSurvey2020a, linFeaturePyramidNetworks2017}. 
Ross Girshick et al.~\cite{RichFeatureHierarchies2014b} introduced the algorithm designed to overcome the problem of selecting large regions during an object detection task. The model performs a selective search on the image by looking for potential objects (region proposals). Therefore, instead of detecting and classifying a larger region, the model divides the image into smaller regions, which ultimately increases the total training time. The Fast R-CNN model~\cite{cazzatoSurvey2020a, girshickFastRCNN2015} was introduced to solve some of the drawbacks of the original model and create a faster model. The two models were approached from the same point of view, with the difference that the input image was fed to the CNN architecture to generate feature maps instead of region proposals, which increased the model's speed and accuracy. The Mask R-CNN model~\cite{cazzatoSurvey2020a, heMaskRCNN2018} was introduced to solve the problem of object detection in images with complex backgrounds, and the model was able to detect objects with high accuracy and segmentation. 

All prior object detection methods focus the item inside the picture using areas. The network does not analyse the picture in its entirety. Instead, the probability-rich regions of the picture that contain the item are the focus. 
You Only Look Once, or YOLO, is an object detection algorithm that differs significantly from the region-based techniques described previously. In YOLO, the bounding boxes and class probabilities for these boxes are predicted by a single neural network. YOLO performs object detection quicker than a conventional object detection algorithms, with speeds ranging from 45 to 145 frames per second. The drawback of the YOLO algorithm is that it struggles to detect small objects in an image, but this changes with more recent versions.

Finally, FPN~\cite{linFeaturePyramidNetworks2017} is not a standalone object detector, and can be classified as a feature extractor that operates in conjunction with object detectors such as R-CNN and Fast R-CNN. The feature extractor accepts a single-scale picture of any arbitrary size as input, and generates correspondingly scaled feature maps at several layers using a fully CNN algorithm. This process is independent of the convolutional architectures' core components. It is a general approach for constructing feature pyramids inside deep convolutional networks for applications such as object identification.

When applying deep learning approaches to problems involving image classification or object detection, one of the most frequent obstacles that arise is a lack of data. Applying data augmentation techniques to a dataset in order to expand its size and variety is a trial-and-error approach to the challenge of dealing with a lack of data~\cite{perez2017}.
The traditional method of data augmentation involves the use of various libraries, such as those described in~\cite{buslaev2020,riba2020}, which provide flexibility and easy-to-use implementation for a variety of augmentations to increase the size and the diversity of the dataset. This approach is known as the ``classic'' method of data augmentation. These libraries include a variety of enhancement methods, including cropping, blurring, colour saturation, contrast, and greyscale scaling, as well as rotation, changing colour channels, and shifting colour~channels.

When it comes to more project specific tasks, the standard data augmentation method cannot generate images that are close to  the preferred real-world data, and it requires a significant amount of time and trial and error to produce the desired results. Therefore, DL models such as Generative Adversarial Networks (GANs), CycleGAN, and U-Nets are the current state-of-the-art methods used to augment datasets and increase their \linebreak size~\cite{Goodfellow2014a, Park2019, ronneberger2015u}. GAN are mainly used to produce synthetic images that follow the same probability distribution as the real images. CycleGAN is a well-known GAN architecture that is typically used to learn image transformations across various patterns, whereas U-Net models focus more on semantic and structural differences between actual and artificial content. These methods is the most advanced currently available.

In addition, the difficulties presented by the underwater environment make the collection of data a laborious job that requires the use of specialised persons and specific equipment. As a consequence of this, it is difficult to build projects that need large underwater datasets. The underwater habitat, the light conditions that are present throughout the picture capturing process, and the task that the image was shot for all play a role in determining the unique problems that come with taking underwater photographs~\cite{Li2017}. When it comes to obtaining data for deep learning models, many researchers have focused primarily on underwater image enhancement and restoration to improve the quality of underwater images~\cite{Corchs2010, Anwar2018, Berman2017}. This is to improve the quality of the images obtained from underwater environments.

The technique of enhancing underwater images has the improvement of the image's visual quality as its goal, and it does not often take into account the physical qualities of light in the water, such as the attenuation coefficient or the light scattering~\cite{Anwar2018}. It is generally agreed that picture enhancement may be implemented far more quickly and is simpler to understand than image restoration. Image restoration is a more sophisticated process that has to take into consideration the physical behaviour of light in water. This is because water reflects light differently than air does. Image restoration requires information on the kind of water present, whether it is coastal or ocean water, as well as the quality of the light propagation in the water~\cite{Corchs2010, Berman2017}. These methods only produce satisfactory outcomes in a controllable underwater environment, and it is difficult to put them into practice in the real world due to the complexity of their implementation and the large number of parameters that need to be taken into consideration~\cite{Corchs2010}.

By including an attenuation coefficient for both the blue--red and blue--green spectrum channels, the technique that was presented by Berman and colleagues~\cite{Berman2017} was able to consider the various light profiles produced by the various underwater settings. The technique they developed is based on the intensity of the image's colour channel at the pixel level; more specifically, the attenuation coefficient incorporates the two spectrum characteristics. In addition to this, the topography of the location, the time of year, and the climate were all taken into consideration. Arnold-Bos and colleagues~\cite{arnold2005} discuss the challenges that the vision of underwater vehicles encounter while operating in underwater conditions and suggest using deconvolution and augmentation approaches. The technique was developed to eliminate light backscattering, the primary feature of noise, and the attenuation inequalities that arise with contrast equalisation. A wavelet filtering approach was then used on the residual picture noise, which may correlate with sensor noise or floating particles. This algorithm helps enhance the edge recognition of underwater images.

To increase the amount and quality of the datasets, more novel techniques as mentioned above, such as data augmentation using GANs, are being employed in various sectors. Some examples of the use of GAN models can be found in the field of neuroscience where, for instance, there is a need to perform segmentation tasks from CT scan images~\cite{sandfort2019}. Additionally, the use of Deep Neural Networks (DNN) and U-Nets to perform segmentation in brain cell representation from Electron Microscopy (EM) images~\cite{ronneberger2015u, fakhry2016} is an example of the application of CycleGAN models for the purpose of data augmentation.

Because of the progress that has been made in DL and CV over the past few years in areas such as image classification, image segmentation, and object detection~\cite{szegedyGoingDeeperConvolutions2014b, Goodfellow2014a, Zhu2020}, there is now an opportunity to develop models that are capable of performing image restoration and image enhancement in a manner that is more accurate and precise. These models have the potential to outperform any of the manual approaches that were used in the past~\cite{Park2019}. Because of the use of Convolution Neural Networks (CNN) and GANs, it is possible, in certain instances, to identify and detect objects with a higher level of accuracy than is possible for humans to attain~\cite{Geirhos2018}.

Zhu et al.~\cite{Zhu2020} proposed a CycleGAN model for image-to-image translation in order to learn the mapping functions between two domain images $X$ and $Y$, translate the domain of the first image based on the second, $G: X \longrightarrow Y$, and vice versa, to translate the second domain based on the first image, $F: Y \longrightarrow X$. This allowed the model to translate the domain of interest. In addition, the authors incorporated two adversarial discriminators, one for the first domain image and one for the second domain image. The purpose of these discriminators was to assess whether or not the output image had been successfully translated to the target domain. In most instances, the results were adequate, and the translation of one image domain to another image domain delivered acceptable output. Nevertheless, the model might become confused between the domains when there is insufficient feature dispersion in the training set.

Currently, DL models are the standard in underwater applications, and the primary emphasis is on picture restoration, image enhancement, and improvement of underwater settings. Anwar et al. suggested using a CNN model that might improve the quality of photographs taken underwater~\cite{Anwar2018}. The network design is composed of convolutional blocks that are all linked to a dense layer at the end. This provides the whole model with modularity. The model's output is an improved picture of the subsea scene, devoid of the cyan and emerald tones present in the original image.

A similar technique for restoring the colours in underwater photographs was used by Chen et al.~\cite{chen2021}, where the authors attempted to reduce the effects of the underwater environment, increase the picture details, and fix the colours in the image. The image incorporates several diverse components, each of which is represented by one of the model's three primary elements. The first part is used to estimate the ambient light of the image; the second part is responsible for the direct transmission estimation, which is a function of both the ambient light and the input image, and the third part is responsible for the reconstruction of the enhanced image.
Li et al.~\cite{liFusionAdversarialUnderwater2019} presented an underwater enhancement method based on GAN models, where they tried to solve underwater degradation effects such as low contrast, colour casts, and haze-like effects using a fusion GAN model on the U45 dataset. The model is utilised by combining the benefits of the inception model architecture~\cite{szegedyGoingDeeperConvolutions2014b} with the deep residual learning framework~\cite{heDeepResidualLearning2015}.

Another research study by {Li et al}.~\cite{liUnderwaterImageEnhancement2020c} approached underwater image enhancement from a different perspective by constructing a large-scale real-world underwater dataset containing 950 images under various light conditions, from natural light to artificial light. The collected data were then tested on the custom Water-Net model to perform image enhancement. 
Furthermore, {Panetta et al.}~\cite{panettaComprehensiveUnderwaterObject2022} went further in underwater object tracking and image enhancement and introduced a benchmark underwater dataset, UOT100. The dataset comprises 104 underwater videos, from which they generated a complete set of 74~K annotated image frames. Additionally, they introduced the CRN-UIE GAN model to perform image enhancement. The model tries to improve underwater object detection performance by correcting the underwater environment's effects. 

Underwater image restoration using real-world images from coral reefs (HICRD) was proposed by {Han et al.}~\cite{hanUnderwaterImageRestoration2021}. They created the custom HICRD dataset to overcome the limitation of previous datasets to capture a more diverse underwater environment. The dataset consists of 9676 images and is used on the Constructive UndeWater Restoration (CWR) model for image restoration. The CWR model at its core utilises GAN models and Representation Learning~\cite{heMomentumContrastUnsupervised2020}, essentially an unsupervised method used to perform image restoration. The CWR model performed satisfactorily, and the end result was close to the reference images without content or structural losses on the generated images.  

In addition, during the last few years, there has been a shift from the conventional techniques toward the substantial use of CNN and GAN models for underwater image repair and enhancement~\cite{Park2019, chen2021, Li2020b, islam2020}. Because of such networks' characteristics, DL models represented a significant advancement in the analysis of underwater photographs. The task of processing and interacting with the underwater world poses a number of difficulties for any autonomous vehicle. DL makes it possible to create more accurate data-driven models of the environment, improving one's capacity to analyse and comprehend that environment. The most notable benefit of DL models is that they can be put into action without the necessity of explicitly describing every facet of the environment and manually coding everything that is required for the operation line by line. This is the distinct advantage that sets them apart from other types of models. DL models can be trained and can learn the most valuable features on their own, provided the necessary data are fed through the network during the training process. As a result, the model would be able to learn the necessary characteristics and parameters for every given job, notwithstanding the complexity of the underwater scenes.

Consequently, as a result of the development of advanced deep learning algorithms, it is now more conceivable than it has ever been to generate underwater photographs that may be as similar as possible to the real world, despite the complexity of such an environment. Because of this, the data collection for marine imagery, which is an essential component of any project relating to the underwater environment, can be made more accessible and will not require direct underwater data, at least in the initial stages of the model development. This will result in savings of time, resources, and funding.

This paper is organised as follows. Section \ref{sec:method} describes the methodology used to generate the underwater photographs. Section \ref{sec:results} describes the results of the model development. Section \ref{sec:conclusions} presents the conclusions of the paper.

\section{Methods for Data Capture and Processing}\label{sec:method}

In the technique that is proposed in this paper, a dataset was compiled based on items that have the potential to be discovered in sub-sea structures, and this was carried out in a laboratory environment. The next step is to compile a dataset that is as accurate a representation as possible of the real-life underwater environment. This is accomplished by utilising common photographs captured under typical atmospheric conditions and image-to-image translation powered by CycleGAN models. Therefore, the production of artificial or “fake” underwater datasets will make it possible to circumvent the challenges associated with the acquisition of real-world underwater photographs.

A primary dataset was compiled by taking pictures of components that are typical of underwater structures. These components include bolts, hex nuts, flanges, anodes, and pipelines. The next step was constructing an expanded data set using traditional Data Augmentation methods such as rotation, cropping, blurring, and changing the colour channel. The CycleGANs model was then trained on the images that were obtained together with additional photographs of the underwater environment using publicly available datasets~\cite{DebrisDatabase,mdjahidulUFO120Datase}. Both datasets, one of which was produced by using data augmentation and the other of which was produced by using CycleGANs (a learning-based method), were compared using the Frechet Inception Distance (FID) metric~\cite{heuselGANsTrainedTwo2018a} in order to determine which method is more appropriate.

\subsection{Formulation of the Proposed Method}

When they were initially presented in the area of DL by Goodfelow et al. in 2014~\cite{Goodfellow2016a}, the Generative Adversarial models and the adversarial loss made a significant contribution in the area of DL. The network consists of two parts: the Generator, which produces pictures that include characteristics of a particular domain, and the Discriminator, which attempts to accurately categorise actual images based on the created ones. Zhu et al.~\cite{Zhu2020} proposed an enhanced adversarial loss in which the loss function employs the least-squared loss of the original loss. This is due to the fact that it demonstrates more stable behaviour during the training process. 

\textls[-15]{Therefore, the adversarial loss can be expressed using Equations (1a) and (1b) as~follows:}

\begin{subequations}
	\begin{equation}\label{eq_uwloss}
		\begin{aligned}
			\pazocal{L}_{GAN_{uw\rightarrow lab}} = \mathbb{E}_{X_{uw}^{real}}[(D_{lab}(G_{lab}(X_{uw}^{real}))-1)^2] \\ 
			+\mathbb{E}_{X_{lab}^{real}}[(D_{lab}(X_{lab}^{real}))^2]
		\end{aligned}
	\end{equation}
	\begin{equation}\label{eq_labloss}
	\begin{aligned}
	  \pazocal{L}_{GAN_{lab\rightarrow uw}} = \mathbb{E}_{X_{lab}^{real}}[(D_{uw}(G_{uw}(X_{lab}^{real}))-1)^2] \\ 
	  +\mathbb{E}_{X_{uw}^{real}}[(D_{uw}(X_{uw}^{real}))^2]
	\end{aligned}
	\end{equation}
	\begin{equation}\label{total_loss}
		\begin{aligned}
		  \pazocal{L}_{GAN}total = \pazocal{L}_{GAN_{lab\rightarrow uw}} +
		  \pazocal{L}_{GAN_{uw\rightarrow lab}}
		\end{aligned}
		\end{equation}
\end{subequations}
where $\mathbb{E}$ is the expected value for the underwater domain $X_{uw}$ and the lab domain $X_{lab}$. The discriminator $D_{uw}$ is responsible for the mapping of the lab images with underwater features and can be expressed as $G_{uw}: X_{uw} \rightarrow X_{lab}$, and discriminator $D_{lab}$ is responsible for the inverse operation $G_{lab}: X_{lab} \rightarrow X_{uw}$. The total adversarial loss $\pazocal{L}_{GAN}total$  is the summation of the lab loss $\pazocal{L}_{GAN_{lab\rightarrow uw}}$  and the underwater loss $\pazocal{L}_{GAN_{uw\rightarrow lab}}$.

\subsection{Cycle Consistency Loss}

The adversarial model has been trained to learn the properties of both the $X_{uw}$ and $X_{lab}$ domains in the cycleGAN models that were presented by Zhu et al.~\cite{Zhu2020}. This indicates that a lab picture may be converted to another image that has characteristics from the area of interest, such as an image of an underwater environment. Because of this, the model has to be able to meet the cycle consistency between the two domains, which can be seen in Equations (2a) and (2b), respectively.

\begin{subequations}
	\begin{align}
	  X_{uw}^{real} \rightarrow G_{lab}(X_{uw}^{real}) \rightarrow G_{uw}(G_{lab}(X_{uw}^{real})) = X_{uw}^{rebuild}\label{ep_uwdomain} \\
	  X_{lab}^{real} \rightarrow G_{uw}(X_{lab}^{real}) \rightarrow G_{lab}(G_{uw}(X_{lab}^{real})) = X_{lab}^{rebuild}\label{ep_labdomain}
	\end{align}
\end{subequations}

Then, the total cycle consistency loss
 $\pazocal{L}_{cycle}^{total}$ is  $\pazocal{L}_{cycle}{uw} + \pazocal{L}_{cycle}{lab}$
where: 
\begin{subequations}
	\begin{align}
	  \pazocal{L}_{cycle}{uw} = \mathbb{E}_{X_{uw}^{real}}[\Vert G_{uw}(G_{lab}(X_{uw}^{real})) - X_{uw}^{real} \Vert_1]\\
	  \pazocal{L}_{cycle}{lab} = \mathbb{E}_{X_{lab}^{real}}[\Vert G_{lab}(G_{uw}(X_{lab}^{real})) - X_{lab}^{real} \Vert_1]
	\end{align}
\end{subequations}

Combining the adversarial and the cycle consistency loss, the total loss of the model will be

\begin{equation} \label{eq1}
	\begin{split}
		\pazocal{L}_{total} = \pazocal{L}_{GAN} + \lambda\pazocal{L}_{cycle}
	\end{split}
\end{equation}
where $\lambda$ is the regularisation hyperparameter factor. The value of $\lambda$ is utilised to regulate and optimise the performance of the network loss~\cite{Goodfellow2016a}.

\begin{figure}[H]
 	\includegraphics[width=13.5cm]{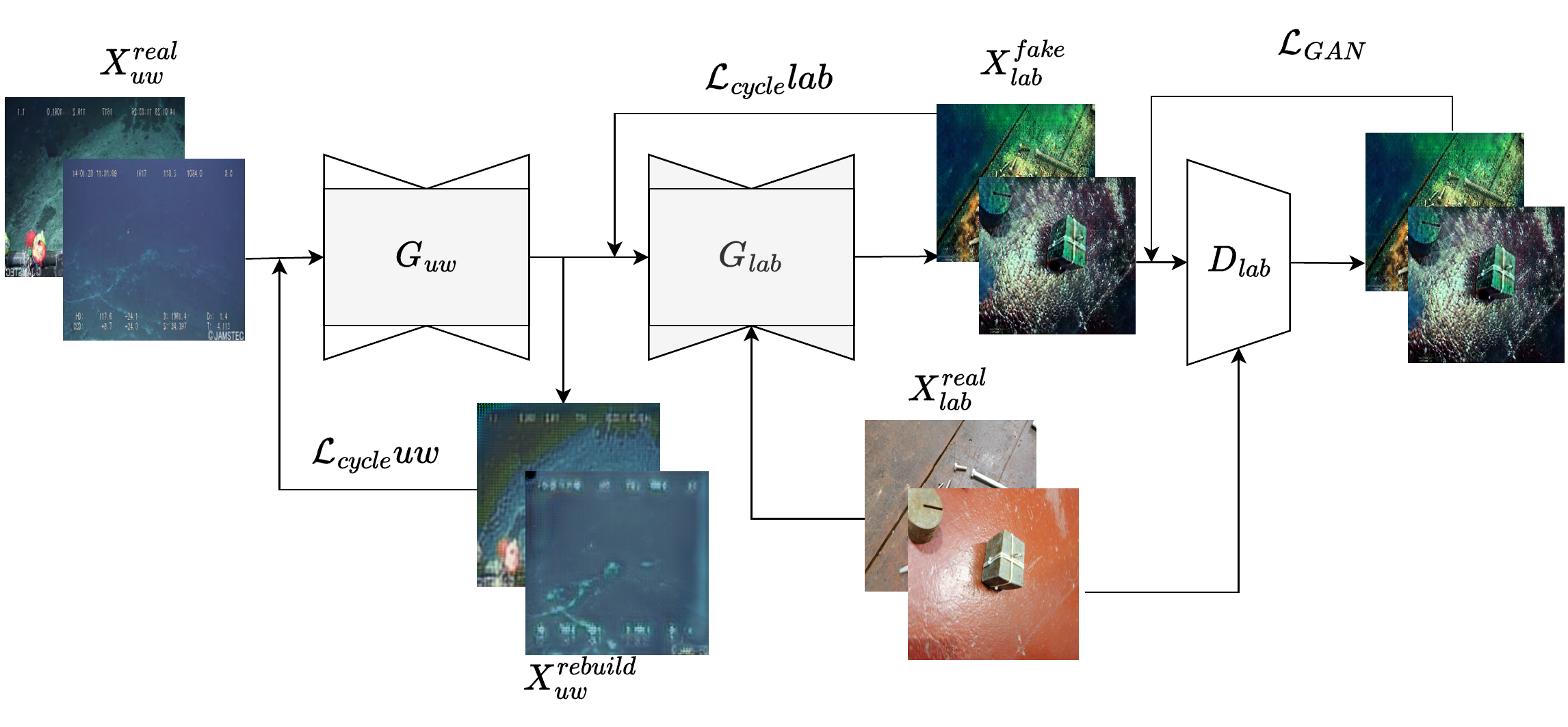}
	\caption{Cyclegan Model for Underwater Data Augmentation.}
	\label{fig:cgan-model}
\end{figure}

The architecture of the UnderWaterCycleGAN (UWCycleGAN) model is shown in \mbox{{Figure} 
 \ref{fig:cgan-model}}. In this model, the image characteristics of the underwater domain $X_{uw}$ are transferred to the required lab image domain $X_{lab}$ using the generator $G_{uw}$. Then, the generator $G_{lab}$ applies the newly acquired features to the original lab pictures $X_{lab}^{real}$, which ultimately leads to the production of the artificial images $X_{lab}^{fake}$. The discriminator $D_{lab}$ is responsible for monitoring and comparing the false pictures with the genuine ones at the very last phase of the network to guarantee that the results are satisfactory. In addition, throughout each step of the process, the model will reconstruct the pictures by making use of the cycle consistency losses. In particular, during the initial step of the process, the model recreates the photographs the underwater environment. $X_{uw}^{rebuild}$ is calculated using the cycle loss $\pazocal{L}_{cycle}uw$. The adversarial loss $\pazocal{L}_{GAN}$ is used to optimise the final image output of the discriminator $D_{lab}$. This is carried out for the fake image $X_{lab}^{fake}$, first by using the $\pazocal{L}_{GAN}$ loss to optimise the output of $G_{lab}$, and then using it to optimise the output of $D_{lab}$.
\vspace{-9pt}

 \subsection{Dataset}\label{data}

Newcastle University's towing tank was used for the data collection needed for the image classification and object identification tasks. In particular, a dataset consisting of five different classes of objects was compiled, including bolts, hex nuts, flanges, pipes, and lead blocks (representing anodes). These objects were chosen because they are common in underwater structures and were readily available during the image data acquisition step. The above objects were initially placed inside the towing tank. Then, videos were captured using an underwater camera. During the acquisition process, different lighting conditions were used to record the underwater objects, including high and low illumination levels (which simulated the underwater environment as much as possible). In addition, to broaden the variety of the dataset, the above objects were used outside the towing tank for image collection. Subsequently, images were extracted from the video files, which resulted in the creation of around 2670 photographs in total.

In addition to the original laboratory dataset, a supplementary 300-image dataset was prepared to serve as the foundation for image classification algorithms. The items in these 300 photographs were manually labelled, and these labelled objects were subsequently removed to generate a dataset including images of five-class object classes. The image classification challenge required only particular items and not the complete scene to be present in the photographs. 
In order to achieve this, a script was created to crop the photographs in the dataset. As inputs, the custom software received the image to be cropped and the file containing the data of the item labels, and it cropped the objects on the provided image. Finally, the script was applied on the prior dataset, resulting in the generation of 2372 object pictures.

\begin{figure}[H]
	\includegraphics[width=8 cm]{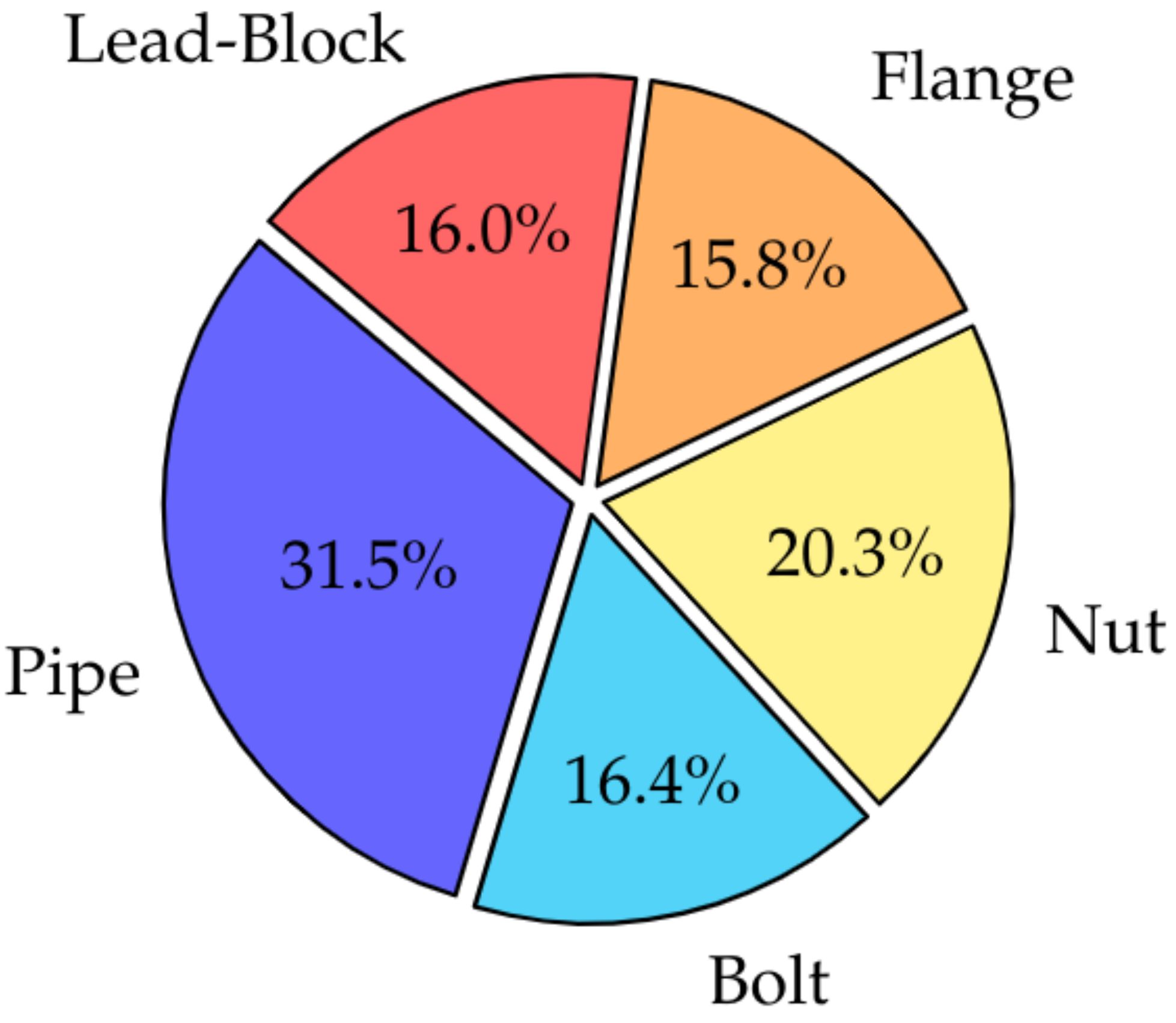}
	\caption{{Object} 
	 Class Distribution.\label{fig:obj-dist}}
\end{figure}

\begin{figure}[H]
\vspace{3pt}
 	\includegraphics[width=0.97\textwidth]{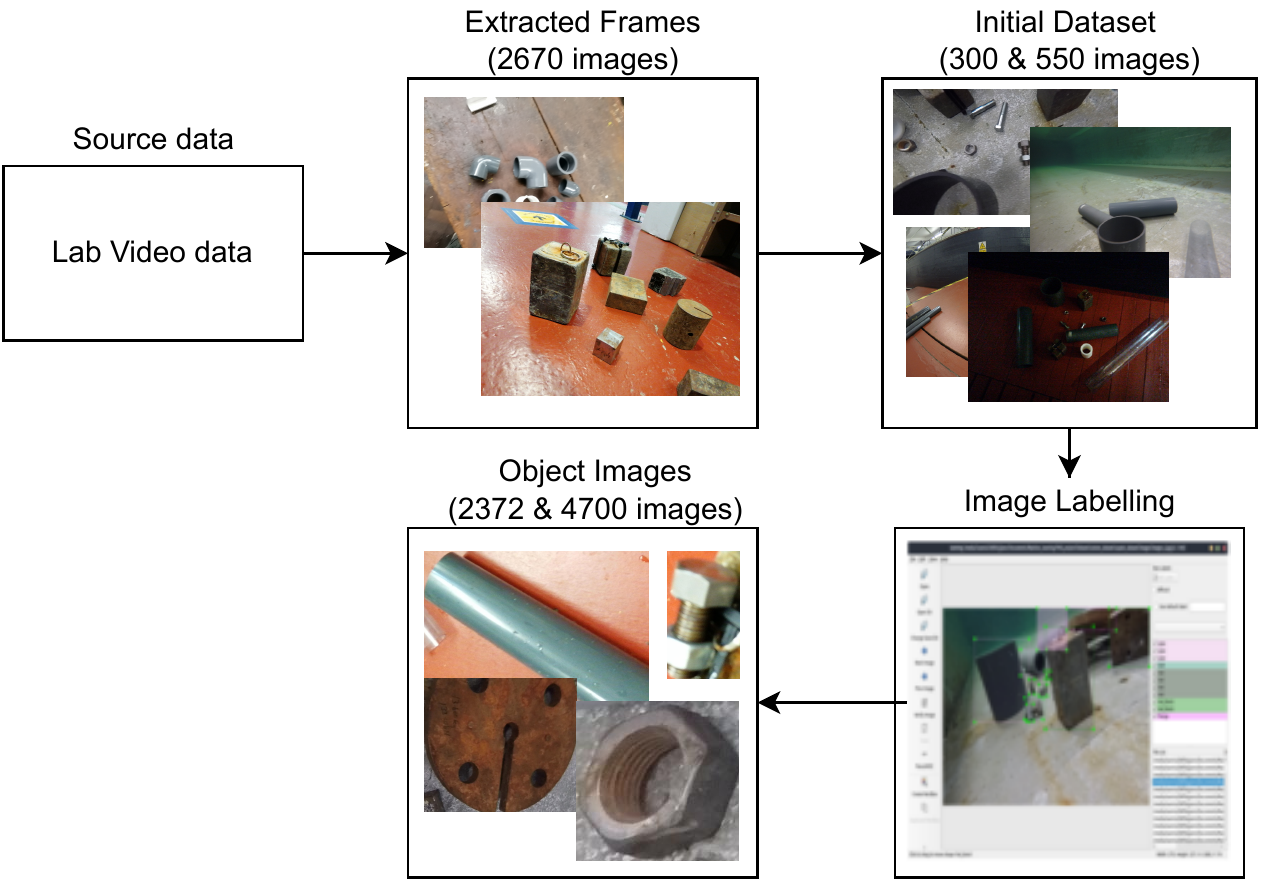}
	\caption{Data Processing Cycle.}
	\label{fig:data-process}
\end{figure}

In order to avoid overfitting and to develop a more robust and accurate image classification model, the original dataset of 300 photographs was increased to 550 images. The same procedure for labelling and extracting the objects from photographs was used, and in the end, the total object dataset had 4700 pictures. The images and objects that were extracted to construct the first and second datasets are shown in Table \ref{tab:table1}, and as was indicated previously, the combination of the two datasets includes 550 individual photographs and 4700 unique class objects. The distribution of the five objects that were used to build the object dataset for the image classification task can be seen in {Figure} \ref{fig:obj-dist}. The steps followed from collecting the video data in the towing tank to the generation of the object images dataset are presented in {Figure} \ref{fig:data-process}. The custom laboratory datasets are available on the ({figshare repository} \url{https://doi.org/10.6084/m9.figshare.20944354.v1}, accessed on 15 August 2022). 

\begin{table}[H] 
\vspace{-4pt}
	\caption{{Towing} 
	 Tank Image Datasets.\label{tab:table1}}
	\newcolumntype{C}{>{\centering\arraybackslash}X}
	\begin{tabularx}{\textwidth}{CCC}
	\toprule
	\textbf{~}	& \textbf{Number of Images}	& \textbf{Number of Objects}\\
	\midrule
	\textbf{First set}  & 300 & 2372 \\ 
		\midrule
		\textbf{Second set} & 250 & 2328 \\ 
		\midrule
		\textbf{Total set}  & 550 & 4700 \\
	\bottomrule
	\end{tabularx}
	\end{table}

\vspace{-12pt}

Last but not least, open source underwater datasets were used to create the dataset that contains the underwater environment. To be more specific, 1500 photos were utilised from the UFO-120 dataset~\cite{mdjahidulUFO120Datase}, while 1170 images were collected from films of the Deep Sea Debris Dataset~\cite{DebrisDatabase} from the Japan Agency for Marine-Earth Science and Technology.

\subsection{Data Augmentation} \label{classdata}
Traditional data augmentation refers to the process of improving visual data using various machine learning frameworks and libraries. In deep learning, and notably in computer vision applications, data augmentation is utilised extensively to enhance the number of datasets as well as the diversity of pictures. For the purpose of this research, the Albumentations~\cite{buslaev2020} library was used to add additional features to the photographs included in the dataset. Albumentations is a well-established library that enables a range of image transformations and augmentation methods. It is specially tailored to work with any of the current Machine Learning frameworks. Therefore, it can be used with almost any of these frameworks.

As a result, the goal was to apply data augmentation in the original dataset, which consisted of 550 images featuring underwater items, to expand the dataset's variety and richness. The enhancement consisted of methods such as rotating and flipping the picture horizontally, increasing the image's saturation, increasing the exposure, and adding noise, as well as transforming it to greyscale. A total of 1980 pictures were produced as a consequence of applying the enhancement techniques to the dataset in a random fashion.

\subsection{Deep Learning Augmentation} \label{dlaug}
The second approach of data augmentation makes use of DL models, especially the use of GAN models. GAN models are essentially two neural networks competing to make more accurate predictions by generating their own training data and automatically detecting and learning patterns to create new samples that plausibly may have been selected from the original dataset. This makes them perfect for data augmentation. A CycleGAN model can take the images that were collected from the towing tank as its first input and an image of an underwater environment as its second input in order to generate outputs that will contain the images that were collected from the towing tank, but with characteristics that are unique to the underwater environment.

The fact that this technique is an unsupervised learning process~\cite{Li2017} means that it may utilise the original dataset without the need to perform any augmentation; nonetheless, it will need photographs that depict the underwater environment in order to function properly. The specifics of the CycleGAN implementation that were utilised to produce the augmented pictures are shown in Figure \ref{fig:cgan-model}.

\begin{figure}[H]
    \includegraphics[width=.23\textwidth]{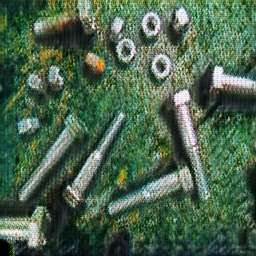}\hfill
    \includegraphics[width=.23\textwidth]{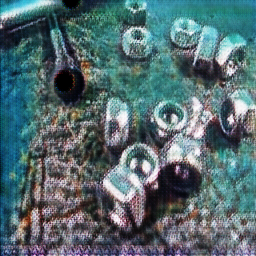}\hfill
    \includegraphics[width=.23\textwidth]{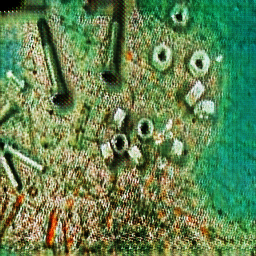}\hfill
    \includegraphics[width=.23\textwidth]{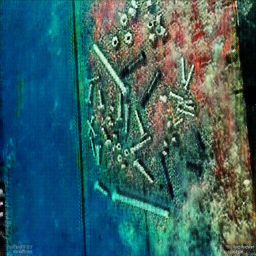}
    \\[\smallskipamount]
    \includegraphics[width=.23\textwidth]{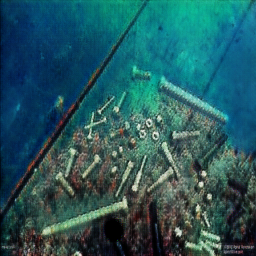}\hfill
    \includegraphics[width=.23\textwidth]{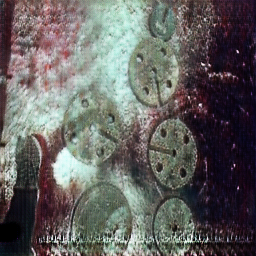}\hfill
    \includegraphics[width=.23\textwidth]{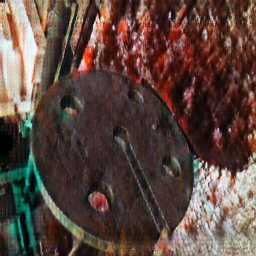}\hfill
    \includegraphics[width=.23\textwidth]{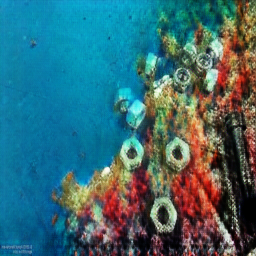}
    \\[\smallskipamount]
    \includegraphics[width=.23\textwidth]{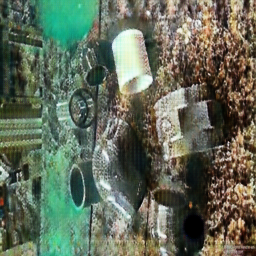}\hfill
    \includegraphics[width=.23\textwidth]{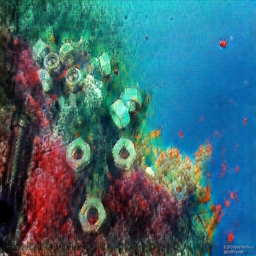}\hfill
    \includegraphics[width=.23\textwidth]{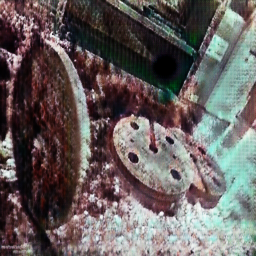}\hfill
    \includegraphics[width=.23\textwidth]{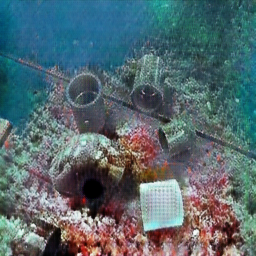}
    \caption{CycleGAN model images output.}\label{fig:cgan-images}
\end{figure}

Figure \ref{fig:cgan-images} illustrates some of the image results that were achieved by employing DL data augmentation. It is evident that these images are far more accurate representations of the subsea environment than the initial data augmentation described in the results section when performing model evaluation with the FID technique. The model requires more optimisation, since in some of the photographs, it might be challenging to identify the individual objects that are there.

\subsection{Image Classification}

The primary aim of the DL data augmentation process is to produce pictures of the type $X_{lab}^{fake}$ that incorporate features from the domain $X_{uw}^{real}$. The last phase involves training an object identification model to determine how well it can identify the items of interest in the data. Before applying the object detection model to the simulated underwater photographs, an image classification model was developed to determine whether or not the five-class objects had been correctly categorised.

\begin{figure}[H]
\vspace{-10pt}
 	\includegraphics[width=\textwidth]{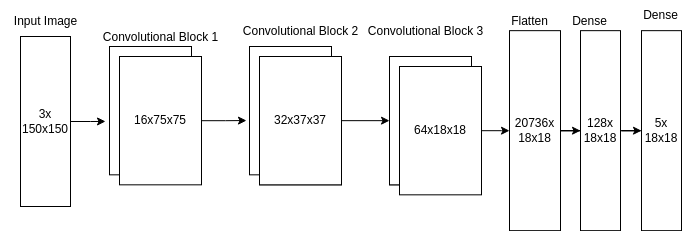}
	\caption{Image Classification model.}
	\label{fig:img-class-md}
\end{figure}

The most essential parameters in the CNN model are listed in Table \ref{tab:cnn-table}. The first step is to crop the input image dimensions to 150 pixels on the horizontal (width, \emph{w}) and vertical (height, \emph{h}) planes, and it contains three colour channels (RGB colour depth). The next step is feature extraction, which happens as the image is processed through the convolutional layers (Conv layer). After the first Conv layer, the channels are increased to 16. After the second Conv layer, they are increased to 32, and after the third Conv layer, they are increased to 64.
Additionally, the Max Pooling technique~\cite{wu2015max} is used after each convolution. This leads to a decrease in the size of the original input, which decreases from $w \times h = 150 \times 50$ pixels to $w \times h = 18 \times 18$ pixels. Next, the output of the Max Pooling layer has to be flattened so that it can appropriately link to the Dense layers, and then lastly, to the classification layer, where the prediction will be made. The details of the image classification CNN model are shown in Figure \ref{fig:img-class-md}.

\vspace{-10pt}

\begin{table}[H] 
\caption{CNN model implementation.\label{tab:cnn-table}}
\newcolumntype{C}{>{\centering\arraybackslash}X}
\begin{tabularx}{\textwidth}{CCC}
\toprule
\textbf{Layer}	& \textbf{Filters}	& \textbf{Output}\\
\midrule
Input image		& 3			& {150} 
 $\times$ 150\\
Convolutional		& 16			&150 $\times$ 150 \\
Max Pooling	& 16			&  75 $\times$ 75  \\
Convolutional		& 32			&75 $\times$ 75   \\
Max Pooling		& 32			&  37 $\times$ 37 \\
Convolutional		& 64			& 37 $\times$ 37  \\
Max Pooling		& 64			& 18 $\times$ 18\\
Flatten		& 20,736			&18 $\times$ 18 \\
Dense		& 128			& 18 $\times$ 18\\
Dense		& 5			&18 $\times$ 18 \\
\midrule
Softmax                             & \multicolumn{2}{c}{Prediction}\\
\bottomrule
\end{tabularx}
\end{table}

\vspace{-2pt}
It is vital to change the original basic CNN model in a manner that will not be prone to overfitting to increase the model's overall accuracy and decrease the amount of loss that occurs during training and validation. This objective may be accomplished by performing several fundamental picture modifications to random images included within the dataset. These transformations include image rotation, image flip, and zoom. After the final Conv layer, an additional layer called a dropout layer is also added. This layer will retain just a subset of the filters or “neurons” and remove the others from the network. The Dropout parameter is now set to 0.2 in this arrangement, which indicates that twenty per cent of the filters will not be used.

Two different models were used during the preliminary training. The first model implemented a basic CNN architecture, consisting of three convolutional layers, and it was used to classify images. The first model was then used in the development of the second model, with the exception of the application of certain fundamental transformations prior to the input to the network and the employment of the dropout layer after the convolution block (Convolutional-Max Pooling). In addition, the dataset used for the CNN model was segmented into training and validation sets, each of which included eighty and twenty percent of the total 4700 object picture dataset, respectively. After this, each model was trained and validated using these two sets.

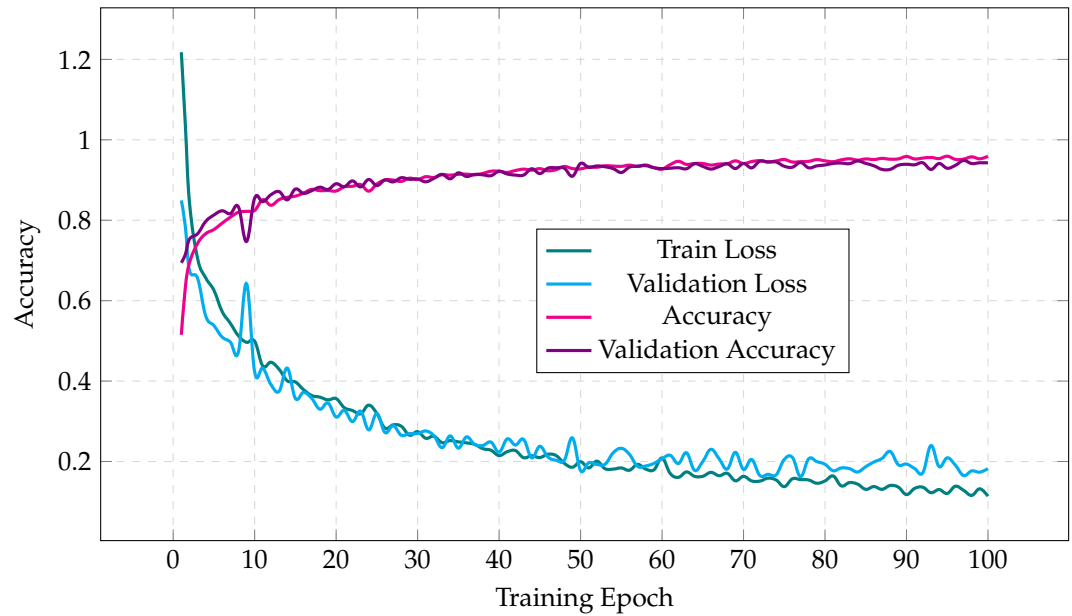
\begin{figure}[H]
\vspace{3pt}
	  \begin{tikzpicture}
		\begin{axis}[
			width=\linewidth,
			height=0.6\textwidth,
			grid=major, 
			grid style={dashed,gray!30},
			xlabel=Training Epoch, 
			ylabel=Accuracy,
			legend style={at={(0.45,0.45)},anchor=west},
		  ]
		  \addplot [color=teal,smooth,tension=0.7,very thick] table[x=Epoch,y=loss,col sep=comma] {data/cnn_4700.csv}; 
		  \addlegendentry{Train Loss}
		  
		  \addplot [color=cyan,smooth,tension=0.7,very thick] table[x=Epoch,y=val_loss,col sep=comma] {data/cnn_4700.csv}; 
		  \addlegendentry{Validation Loss}
		  
		  \addplot [color=magenta,smooth,tension=0.7,very thick] table[x=Epoch,y=accuracy,col sep=comma] {data/cnn_4700.csv}; 
		  \addlegendentry{Accuracy}
		  
		  \addplot [color=violet,smooth,tension=0.7,very thick] table[x=Epoch,y=val_accuracy,col sep=comma] {data/cnn_4700.csv}; 
		  \addlegendentry{Validation Accuracy}
		  
		\end{axis}
	  \end{tikzpicture}
	  \caption{CNN model results trained on 4700 images.}
	  \label{fig:scnn-r-3}
  \end{figure}

The results of running the first CNN model on the initial object dataset consisting of 4700 pictures are shown in Figure \ref{fig:scnn-r-3}. The model structure may be seen in Table \ref{tab:cnn-table}. The model's accuracy improves at a comparable pace over the whole training period of \mbox{$100$ epochs} in both the training and validation sets; however, the validation accuracy is somewhat lower than the training accuracy. Training loss and validation loss follow the same diminishing trend, with the validation loss starting a little higher than the \mbox{training~loss}.

\vspace{-7pt}


%

\subsection{Object Detection}
Object detection models were used in the last phase to identify the five different class items. Initially, the YOLO v1 algorithm~\cite{Redmon2016b} was selected, but this was later upgraded to the YOLO v4 algorithm~\cite{Bochkovskiy2020}. Based on the original paper~\cite{Redmon2016b}, an implementation of the YOLO v1 model was selected primarily due to its speed during the training of the model, the simplicity of the model architecture, which involved a single forward CNN architecture, and lastly, its efficiency, because it involves fewer detection bounding boxes while training in comparison to other more demanding models such as R-CNN models~\cite{RichFeatureHierarchies2014b}. The YOLO v4 model was then used to produce more accurate detection results on the dataset at shorter training times and with greater hardware utilisation. The utilised code is publicly accessible from the authors~\cite{bochkovskiyYoloV4V32021}.

The training and assessment of all models were carried out on a double NVIDIA Tesla V100 GPU using the Rocket High-Performance Computing (HPC) Service offered by Newcastle University.

\section{Results}\label{sec:results}
The approach used in this study to construct a bespoke dataset comprised of lab photographs and including underwater features has been outlined in Section \ref{sec:method}. Essentially, the method takes an input picture called $X_{uw}^{real}$ and then passes this image on to the generator $G_{uw}$ in order to extract the features, and finally rebuilds the image called $X_{uw}^{rebuild}$. The extracted features, together with the input image $X_{lab}^{real}$ are then used in the generator $G_{lab}$ to generate $X_{uw}^{fake}$, which is then assessed by the discriminator $D_{lab}$. Lastly, the extracted features are evaluated by the discriminator $D_{lab}$. In addition, the outputs of the models are applied to the image classification model, as discussed.

The UWCycleGAN model was trained on both the full-scale and object image datasets. The training of the model was performed on the Rocket HPC. Some key experimental details include the size of each dataset, the image size, the training time required to complete the task, the batch size, and the learning rate of the model. Table \ref{tab:exp-details} gives a summary of those experimental details. For the full-scale image dataset, the dimensions of the input images were $256 \times 256$ pixels with batch size $4$ images per batch and the learning rate was $2 \times 10^{-4}$. The model training was performed on an NVIDIA Tesla V100 GPU, consuming 14.2 GB of RAM for 18 h. Similarly, the object image dataset was trained on the same hardware, but the input image was set to $128 \times 128$ pixels with a batch size of $8$ images per batch and a learning rate of $2 \times 10^{-4}$, consuming 15 GB RAM for 22 h of training. Additionally, different image transformations were used in the model during the preprocessing phase to allow the model to generalise better in an unseen dataset. Such transformations include Gaussian Blur, Horizontal Flip, and Random Rotation.

\begin{table}[H] 
\vspace{-2pt}
\caption{UWCycleGAN model experimental details.\label{tab:exp-details}}

\begin{adjustwidth}{-\extralength}{0cm}
\newcolumntype{C}{>{\centering\arraybackslash}X}
\begin{tabularx}{\fulllength}{m{1.9cm}<{\raggedright}m{1.8cm}<{\raggedright}m{1.8cm}<{\raggedright}m{1.7cm}<{\raggedright}m{2cm}<{\raggedright}m{2.2cm}<{\raggedright}m{1.8cm}<{\raggedright}m{2.2cm}<{\raggedright}}
\toprule
\textbf{CycleGAN }	& \textbf{Dataset Size}	& \textbf{Image Size}&\textbf{Hardware}	& \textbf{RAM Usage}	& \textbf{Training Time}&\textbf{Batch Size}	& \textbf{Learning Rate}	\\
\midrule
Full Image		& 2670  			& 256 × 256  & Tesla V100 & 14.2 GB                                              & 18 h                                                   & 4                                                     & $2 \times 10^{-4}$\\
Object Image		& 4700			&128 × 128  & Tesla V100 & 15.0 GB                                              & 22 h                                                   & 8                                                     & $2 \times 10^{-4}$\\
\bottomrule
\end{tabularx}
\end{adjustwidth}
\end{table}


\subsection{Model Evaluation} \label{model-evaluation}
The end output of an adversarial model is an image and, more specifically for the purpose of this study, the result of the UWCycleGAN model is the false image $X_{lab}^{fake}$. Because of this, the generated output image $X_{lab}^{fake}$ needs to be compared to the real input image $X_{real}$ in order to determine whether the artificial image is an accurate representation of the underwater domain. The FID technique, which is the metric that produces the distance between two feature vectors of the $X_{lab}^{fake}$ and $X_{real}$, is the most effective approach to achieve such an analysis~\cite{brownleeHowImplementFrechet2019}.

%
%
%
%
%
%

The FID metric is described by the following equations:
\begin{equation}\label{eq_fid}
	\begin{aligned}
	  \delta^2(\mu_{1,2}, C_{1,2}) = M_{1,2} + \tau_{r}(C_{1,2})
	\end{aligned}
\end{equation}
\begin{equation}
	\begin{aligned}
		M_{1,2} = \|\mu_1 - \mu_2\|_2^2 
	\end{aligned}
\end{equation}
\begin{equation}
	\begin{aligned}
		C_{1,2} = (C_1 + C_2 - 2\sqrt(C1 * C2))
	\end{aligned}
\end{equation}
where $\delta$ is the Frechet distance, which is also known as the Wasserstein-2 distance~\cite{heuselGANsTrainedTwo2018a}; $\mu_1$ and $\mu_2$ are the feature-wise mean of the real and the fake output images. $C_1$ and $C_2$ are the covariance matrices of the real and generated images, and $\tau_r$ is the trace linear operation of the square matrices. The FID criterion should be as low as practicable and, therefore, should be zero in the case of identical photos. In addition, the assessment of the Frechet distance is predicated on the first implementation in~\cite{heuselGANsTrainedTwo2018a}, which was carried out using the PyTorch ML framework~\cite{Seitzer2020FID}.

The outcomes of the UWCycleGAN model are shown in Table \ref{tab:table2}. The UWCycleGAN model was tested with two distinct varieties of images: the first test was performed on the dataset consisting of the original 2670 images that were gathered in the laboratory, and the second test was carried out on the dataset consisting of the 4700 objects that were extracted from those original images. Finally, the FID evaluation was carried out on the images produced by the classical data augmentation. Given that the augmentation is carried out directly on the dataset, the only metrics that can be compared are those that compare \mbox{$X_{uw}^{real}$ vs. $X_{uw}^{fake}$} and $X_{lab}^{real}$ vs. $X_{lab}^{fake}$, respectively.

\begin{table}[H]
\vspace{-4pt}
\tablesize{\fontsize{7}{7}\selectfont}
\caption{{FID} 
 scores.\label{tab:table2}}
\setlength{\cellWidtha}{\fulllength/8-2\tabcolsep+0.4in}
\setlength{\cellWidthb}{\fulllength/8-2\tabcolsep-0.1in}
\setlength{\cellWidthc}{\fulllength/8-2\tabcolsep-0.1in}
\setlength{\cellWidthd}{\fulllength/8-2\tabcolsep-0.1in}
\setlength{\cellWidthe}{\fulllength/8-2\tabcolsep-0in}
\setlength{\cellWidthf}{\fulllength/8-2\tabcolsep-0in}
\setlength{\cellWidthg}{\fulllength/8-2\tabcolsep-0in}
\setlength{\cellWidthh}{\fulllength/8-2\tabcolsep-0.1in}

	\begin{adjustwidth}{-\extralength}{0cm}
		\begin{tabularx}{\fulllength}{>{\centering\arraybackslash}m{\cellWidtha}>{\centering\arraybackslash}m{\cellWidthb}>{\centering\arraybackslash}m{\cellWidthc}>{\centering\arraybackslash}m{\cellWidthd}>{\centering\arraybackslash}m{\cellWidthe}>{\centering\arraybackslash}m{\cellWidthf}>{\centering\arraybackslash}m{\cellWidthg}>{\centering\arraybackslash}m{\cellWidthh}}

			\toprule
			\textbf{FID Score}	& \textbf{~}	& \boldmath{$X_{uw}^{real}$} \textbf{vs.} \boldmath{$X_{lab}^{real}$}     & \boldmath{$X_{uw}^{real}$} \textbf{vs.} \boldmath{$X_{uw}^{fake}$}&\textbf{$X_{lab}^{real}$} \textbf{vs.} \boldmath{$X_{lab}^{fake}$}	& \boldmath{$X_{uw}^{real}$} \textbf{vs.}\boldmath{ $X_{uw}^{rebuild}$}	& \boldmath{$X_{lab}^{real}$} \textbf{vs.}\boldmath{ $X_{lab}^{rebuild}$}     & \boldmath{\boldmath{$X_{real}$} \textbf{vs.} \boldmath{$X_{lab}^{fake}$}}\\
			\midrule
			\multirow{2}{*}{UWCycleGAN}		& Full Image   & 301.05 & 21.92  & 20.31  & 11.93 & 7.38  & \textbf{42.12}\\
	~&	 Object Image & 326.10 & 29.19  & 36.79  & 20.58 & 15.90 & - \\
			\mbox{Classical~Augnentation}		& -            & -      & 174.26 & 212.91 & -     & -     & - \\
			\bottomrule
		\end{tabularx}
	\end{adjustwidth}
\end{table}

\vspace{-4pt}
As stated previously, a lower value of the FID metric implies that the images are more comparable to one another. The FID value is high for images that are entirely distinct from one another, denoted by the notation $X_{uw}^{real}$ and $X_{lab}^{real}$, respectively, as shown in Table~\ref{tab:exp-details}, but it is much lower for images that are comparable to one another. The score may drop as low as 7.38 when comparing $X_{lab}^{real}$ vs. $X_{lab}^{rebuild}$ and $X_{uw}^{real}$ vs. $X_{uw}^{rebuild}$, respectively. Each picture that was processed using the cycleGAN model has a counterpart that was stored in the object dataset. However, the score for each comparison is higher because the object dataset contains images that include only the area of interest (pipes, flanges, etc.). This makes it significantly more difficult for the model to perform as satisfactorily as it did in the first scenario, which used the entire image.

The most important comparison is $X_{lab}^{real}$ and $X_{lab}^{fake}$, where the FID score is 20.31. This is because the goal is to make images that include underwater features. Given this score, it can be deduced that the fake lab underwater picture is quite similar to the real underwater image. Equivalent results were obtained with the object dataset; however, with the classical data augmentation, the values are significantly different due to the absence of underwater features in the images, which instead make use of various transformations such as hue transformation, image blur, noise, and saturation.

\begin{figure}[H]
\vspace{-4pt}
\hspace{-8pt}
    \begin{tikzpicture}
        \matrix[matrix of nodes]{
        \includegraphics[width=2.5cm, height=2.5cm]{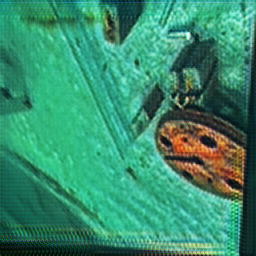} &
		\includegraphics[width=2.5cm, height=2.5cm]{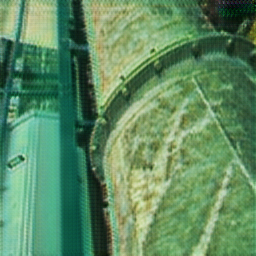} & 
        \includegraphics[width=2.5cm, height=2.5cm]{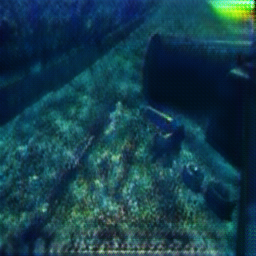}&
		\includegraphics[width=2.5cm, height=2.5cm]{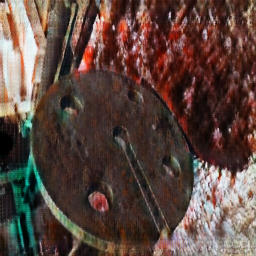} & 
		\includegraphics[width=2.5cm, height=2.5cm]{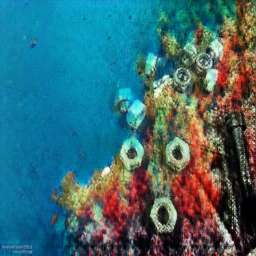}\\
        \includegraphics[width=2.5cm, height=2.5cm]{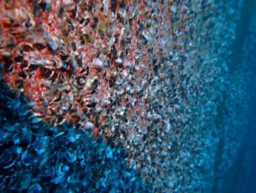} & 
		\includegraphics[width=2.5cm, height=2.5cm]{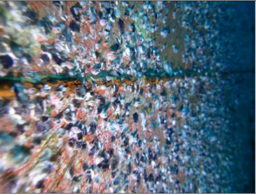} & 
		\includegraphics[width=2.5cm, height=2.5cm]{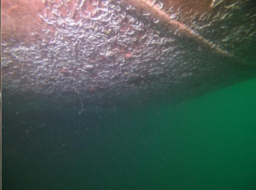} & 
		\includegraphics[width=2.5cm, height=2.5cm]{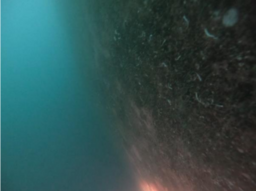} &
		\includegraphics[width=2.5cm, height=2.5cm]{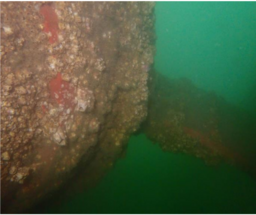}\\
        };
    \end{tikzpicture}
    \caption{Real and artificial underwater images.}
	\label{fig:real_uw_img}
\end{figure}

Additionally, to test the artificial underwater pictures that the UWCycleGAN model produced, actual underwater photographs were employed. These pictures were taken from reports of ship hull cleaning processes that were found on MAST Maritime Services' website~\cite{mastUnderwaterServicesMAST2021}. The photos utilised for the FID assessment are shown in Figure \ref{fig:real_uw_img}. The photographs of the ship's hull before the cleaning procedure are shown in the top row, while those produced by the UWCycleGAN model are displayed in the bottom row. The analysis of these pictures can be seen in Table~\ref{tab:exp-details}, which presents a side-by-side comparison of the underwater picture taken in the actual world and the one created in a lab. The FID score of 20.31 for $X_{lab}^{real}$ vs. $X_{lab}^{fake}$ is  lower than the score of 42.12 when comparing $X_{lab}^{fake}$ and $X_{real}$. This discrepancy is to be expected given that the underwater photographs were taken in the actual subsea environment and only show the unclean portion of the ship hull, which does not include any items such as those found in the laboratory data. As a result, these pictures do not have the same level of detail as the lab items, resulting in lower ratings.

\subsection{Object Detection}
After the image classification model had produced the anticipated results, it was necessary to continue to the next step, which was the training of the object detection model. As discussed in Section \ref{sec:method}, YOLO v1 and YOLO v4 were the models that were~selected.

\subsubsection{YOLO v1}
In order for the YOLO v1 algorithm to function appropriately on the towing tank dataset, the first version of the method had to be updated. It was necessary to change the parameters for the input classes because the initial model was trained on the Pascal VOC dataset, which contains 20 classes~\cite{Redmon2016b}, but the towing tank dataset only has five, and the dataset that was used for the object detection model was the initial towing tank dataset, which contains 550 images. The object identification process requires a significant amount of computing resources due to the fact that the algorithm must pinpoint several items inside a picture. Hence, the Rocket HPC was used for this phase of the project. The \mbox{YOLO v1} was trained on a single NVIDIA Tesla V100 GPU, and it took approximately \mbox{four hours} to complete the task.

\subsubsection{YOLO v4}
\begin{figure}[H]
\vspace{-4pt}
    
        \includegraphics[width=6.5cm, height=6.5cm]{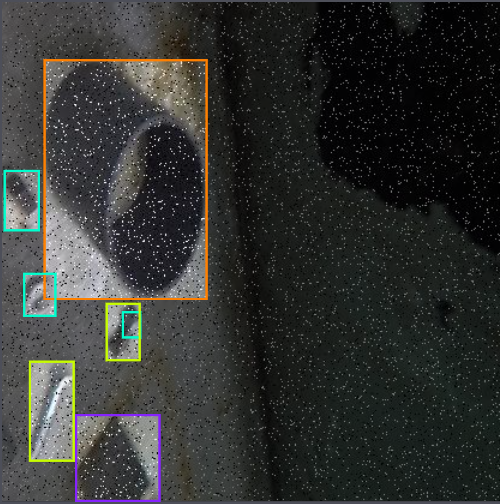}
		\includegraphics[width=6.5cm, height=6.5cm]{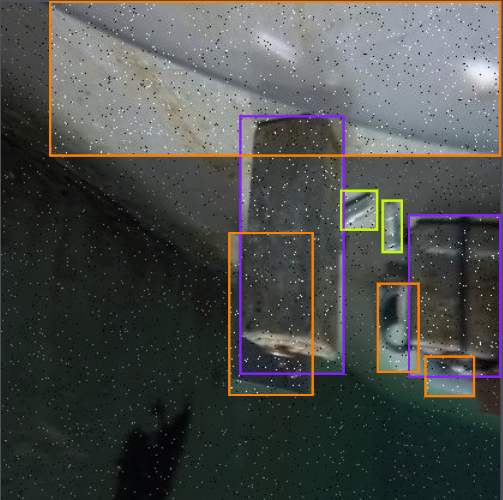}

    \caption{YOLO v4 Dataset preprocessing.}
	\label{fig:yolo-v4-noise}
\end{figure}

The aim is to develop computer vision models that apply to as many real-world scenarios as possible, and this can be achieved by vigorously training the model and allowing it to generalise, which will allow it to use unseen data. 
Building deep learning models that predict an environment that may not have prior knowledge, particularly the underwater environment, is quite challenging. 
Training a DL model with a limited dataset may lead to overfitting, and hence poor results compared to a larger image dataset. Given the uneven or sparse sampling of points in the high-dimensional input data space, small datasets may also provide a more difficult mapping challenge for neural networks to learn. Adding noise to inputs during training is one way to make the input space smoother and easier~\cite{vincentStacked2010, goodfellowDeepLearning2016}. Figure \ref{fig:yolo-v4-noise} shows the effects of the salt-and-pepper noise on the training dataset.
During the training of the YOLO v4 object detection model, the following augmentation techniques were applied:

\begin{itemize}
	\item 50\% probability of horizontal flip.
	\item Random Gaussian blur between 0 and 1.25 pixels.
	\item Salt-and-pepper noise was applied to 8 per cent of pixels.
\end{itemize}

The amount of data input into an object detection model is significant, and this results in the need for computing capacity, such as graphics processing units (GPUs), to reduce the training time. Even if YOLO v1 achieved good results and can detect the objects it was trained on, it still needs to be utilised for the object detection task to be as efficient and effective as possible and reduce the training time as much as possible. To achieve better model performance, the algorithm needs to function properly on the available hardware. The most reliable option to solve this problem is to use an already optimised detection algorithm such as the YOLO v4 model. This model delivers more accurate results for the object identification tasks, and shorter training periods. In comparison with the initial YOLO v1 training time, the YOLO v4 was able to achieve the same results in one hour. The YOLO v4 model that was used for the training was based on the official model~\cite{Bochkovskiy2020}. Experimental training details of {YOLO v1} 
 and {YOLO v4} are shown in Table \ref{tab:yolo-exp-details}.

\begin{table}[H] 
\caption{YOLO object detection model experimental details.\label{tab:yolo-exp-details}}

\begin{adjustwidth}{-\extralength}{0cm}
\newcolumntype{C}{>{\centering\arraybackslash}X}
\begin{tabularx}{\fulllength}{m{2.5cm}<{\raggedright}m{1.5cm}<{\raggedright}m{2cm}<{\raggedright}m{2cm}<{\raggedright}m{2cm}<{\raggedright}m{2cm}<{\raggedright}m{1.5cm}<{\raggedright}m{2.5cm}<{\raggedright}}
\toprule
\textbf{Object Detection}	& \textbf{Dataset Size}	& \textbf{Image Size}&\textbf{Hardware}	& \textbf{RAM Usage}	& \textbf{Training Time}&\textbf{Batch Size}	& \textbf{Learning Rate}	\\
\midrule
YOLO v1 & 1128 & 416 × 416 & Tesla V100 & 4.2 GB & 3.8 {h} & 16 & 0.01\\
	YOLO v4 & 1128 & 416 × 416 & Tesla V100 & 2.1 GB & 1 {h} & 16 & 0.01 \\
\bottomrule
\end{tabularx}
\end{adjustwidth}
\end{table}

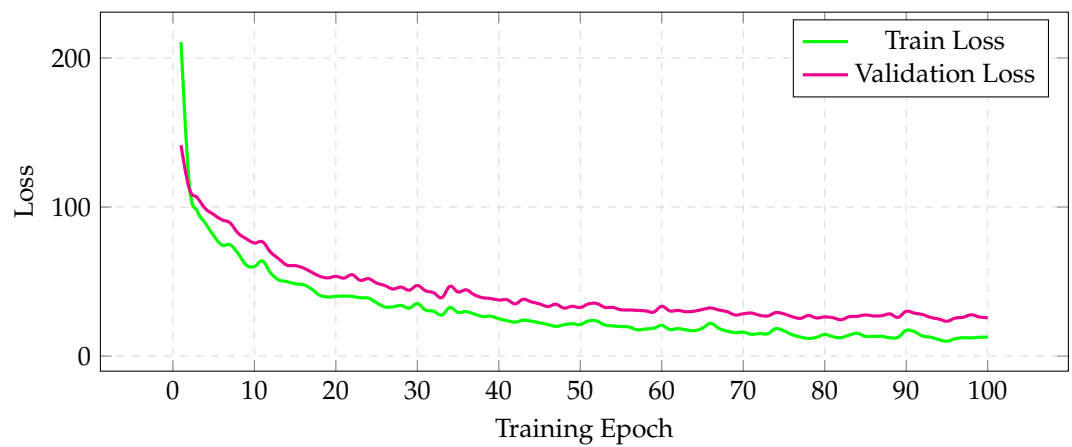
\begin{figure}[H]
	  \begin{tikzpicture}
		\begin{axis}[
			width=\linewidth, 
			height=0.44\textwidth,
			grid=major, 
			grid style={dashed,gray!30},
			xlabel=Training Epoch, 
			ylabel=Loss,
			legend style={anchor=north east},
		  ]
		  \addplot [color=green,smooth,tension=0.7,very thick] table[x=Epoch,y=train_loss,col sep=comma] {data/yolov4.csv}; 
		  \addlegendentry{Train Loss}
		  
		  \addplot [color=magenta,smooth,tension=0.7,very thick]table[x=Epoch,y=val_loss,col sep=comma] {data/yolov4.csv}; 
		  \addlegendentry{Validation Loss}
		  
		\end{axis}
	  \end{tikzpicture}
	  \caption{YOLO V4 loss.}
	  \label{fig:yolo-tvl}
  \end{figure}

The model was altered to function based on the five classes; also, parameters such as the batch number for the GPU that was utilised and the picture input size were updated to conform to the model's needs. Figure \ref{fig:yolo-tvl} shows the model's outcomes in terms of training and validation Loss. As can be seen from the plot, both losses follow the same decreasing trend, and the model converges with just a minor difference between them. In a perfect world, the training Loss and validation Loss should be equal; if they are not, this is an indicator of some overfitting. The gap between them remains the same after approximately epoch 50, and there is no indication that the validation loss will deviate further compared to the training Loss. Nevertheless, it is impossible to achieve ideal results, and some overfitting is always acceptable.

\begin{figure}[H]
\vspace{-6pt}
    
        \includegraphics[width=\textwidth, height=6.5cm]{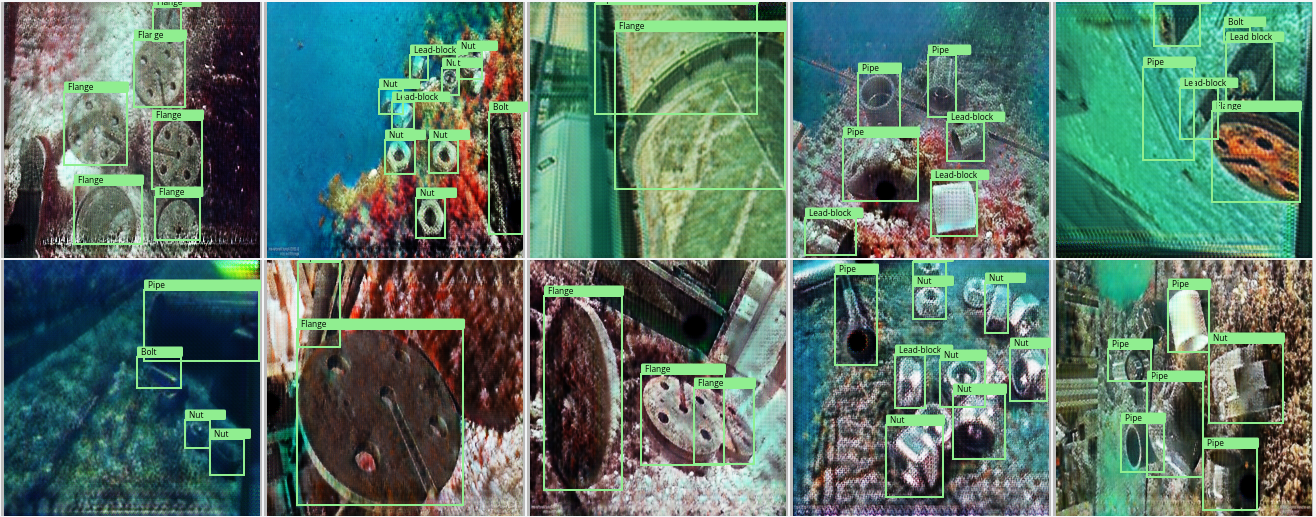}

    \caption{YOLO object detection.}
	\label{fig:yolo-v4-obj}
\end{figure}

Finally, the inference YOLO v4 model, which was trained on the laboratory dataset, was used to validate the UWCycleGAN model's artificial images. Figure \ref{fig:yolo-v4-obj} presents the results of the detection model. The model is able to identify the various objects in the majority of the photographs; nevertheless, there are some instances in which it misclassifies the objects in the image; for example, in the first image, the model identifies the pipe as a lead block. The model needs to be trained in a larger and more diverse dataset to achieve better results and increased accuracy. The model's performance shows that it is able to detect the objects in the images, and it can be used to identify the objects in the artificial images generated by the UWCycleGAN model. However, future work focused on collecting relevant real-world underwater data to use for the training of the object detection model should be conducted. 

\section{Conclusions}\label{sec:conclusions}
Inspection and maintenance of underwater structures are among the numerous underwater-related applications that encounter difficulties due to a lack of data. Researchers are tackling the issue using new technologies made available by the rapid developments in the field of artificial intelligence in recent years. However, they have primarily focused on aspects of image enhancement or restoration. This is because of the difficulties that have arisen as a result of the lack of publicly available datasets from underwater environments. This paper attempts to solve the problem of easily accessible underwater datasets in light of the growing complexity of the underwater world and the absence of a tool to produce artificial underwater photos.

The main contribution of the present work is the development of a DL model used to generate images with underwater features. The model uses images of objects that can be found in underwater structures, such as pipes, anodes, flanges, bolts and nuts, taken under lab conditions, and images containing underwater scenes taken from public datasets. 

As the research has demonstrated, it is clear that the UWCycleGAN model can be used to generate images with underwater features. Furthermore, the underwater domain of the artificially generated images was validated against real-world underwater images of the underwater ship hull section. The results between the generated and real-world images shown that the model can generate realistic underwater features. Finally, to investigate if the artificial underwater images can be used for object detection, they were tested on the initial YOLO v4 object detection model and the results show that the model can generalise satisfactorily and detect the objects.  
Since object detection is crucial  for modern underwater operations such as underwater structural inspection and maintenance, the proposed method can be used to create rapidly datasets containing the desired objects and features to test the initial performance of such applications. 
Although it is possible to produce synthetic images of the underwater environment only using custom images and perform object detection with good results, to complete validation of the method proposed in this paper, new real-world underwater images from offshore structures need to be collected. The two real-world and artificial datasets will be compared for similarities using the FID metric introduced in Section \ref{model-evaluation}, and then the YOLO v4 model will be trained on the artificial dataset and tested on the real-world dataset.

\vspace{6pt}

\authorcontributions{Conceptualisation, I.P., M.H., R.N. and D.T.; methodology, I.P., M.H., R.N. and D.T.; software, I.P.; validation, I.P.; formal analysis, I.P.; investigation, I.P.; resources, I.P.; data curation, I.P.; writing---original draft preparation, I.P.; writing---review and editing, I.P., M.H., R.N. and D.T.; visualization, I.P.; supervision, M.H., R.N. and D.T.; project administration,  I.P., M.H., R.N. and D.T.; funding acquisition,  I.P., M.H., R.N. and D.T. All authors have read and agreed to the published version of the manuscript.}

\funding{\textls[-15]{This research was funded by EPSRC Doctoral Training Programme grant number EPN5095281}.}

\institutionalreview{Not applicable.}

\informedconsent{Not applicable.}


\dataavailability{\textls[-15]{The datasets for the UWCycleGAN, and YOLO v4 models are \mbox{available at}}: \url{https://doi.org/10.6084/m9.figshare.20944354.v1}, accessed on 15 August 2022.} 

\acknowledgments{\textls[-15]{The authors would like to express our gratitude to the staff of Hydrodynamics Lab at
Newcastle University for their support and help during the image acquisition in the towing tank}.}

\conflictsofinterest{The authors declare no conflict of interest.}

\abbreviations{Abbreviations}{
The following abbreviations are used in this manuscript:\\

\noindent 
\begin{tabular}{@{}m{2.2cm}<{\raggedright}m{12cm}<{\raggedright}}
    AI & Artificial Intelligence.\\
    AUVs & Autonomous Underwater Vehicles.\\
    \end{tabular}
    
    \noindent 
\begin{tabular}{@{}m{2.2cm}<{\raggedright}m{12cm}<{\raggedright}}
    CNN & Convolutional Neural Network.\\
    CV & Computer Vision.\\
    DL & Deep Learning.\\
    DNN & Deep Neural Networks.\\
    EM & Electron Microscopy.\\
    GANs & Generative Adversarial Networks.\\
    HPC & High Performance Computing.\\
    ROVs & Remote Operated Vehicles.\\
    UWCycleGAN & UnderWaterCycleGAN.\\
	$X_{uw}^{real}$ & Real-world underwater environment image.\\
    $X_{uw}^{rebuild}$ & Reconstructed real world underwater environment image.\\
    $X_{lab}^{real}$ & Laboratory image.\\
    $X_{lab}^{fake}$ & Artificial underwater image. \\
    $D_{uw}$ & Underwater environment generator. \\
    $D_{lab}$ & Laboratory environment generator. \\ 
\end{tabular}
}


\appendixtitles{no} 



\begin{adjustwidth}{-\extralength}{0cm}

\reftitle{References}

\end{adjustwidth}
\end{document}